\documentclass[letterpaper]{article} % DO NOT CHANGE THIS
\usepackage{aaai23}  % DO NOT CHANGE THIS
\usepackage{times}  % DO NOT CHANGE THIS
\usepackage{helvet}  % DO NOT CHANGE THIS
\usepackage{courier}  % DO NOT CHANGE THIS
\usepackage[hyphens]{url}  % DO NOT CHANGE THIS
\usepackage{graphicx} % DO NOT CHANGE THIS
\usepackage{natbib}  % DO NOT CHANGE THIS AND DO NOT ADD ANY OPTIONS TO IT
\usepackage{caption} % DO NOT CHANGE THIS AND DO NOT ADD ANY OPTIONS TO IT
\pdfoutput=1%For ARXIV

\usepackage{algorithm}
\usepackage{algorithmic}
\usepackage[switch]{lineno}
\usepackage{amsmath}
\usepackage{subfig}
\usepackage{amssymb}
\usepackage{booktabs}
\usepackage{multirow}
\usepackage{multicol}
\usepackage{tikz}
\newcommand*{\circled}[1]{\lower.7ex\hbox{\tikz\draw (0pt, 0pt)%
    circle (.5em) node {\makebox[1em][c]{\small #1}};}}
\usepackage{array}

\definecolor{cGreen}{RGB}{47,139,69}
\definecolor{Red}{RGB}{255,0,0}
    
\usepackage{newfloat}
\usepackage{listings}
\DeclareCaptionStyle{ruled}{labelfont=normalfont,labelsep=colon,strut=off} % DO NOT CHANGE THIS
\floatstyle{ruled}
\newfloat{listing}{tb}{lst}{}
\floatname{listing}{Listing}
\title{Compact Transformer Tracker with Correlative Masked Modeling}

\author{
    Zikai Song\textsuperscript{\rm 1}, 
    Run Luo\textsuperscript{\rm 1}, 
    Junqing Yu\textsuperscript{\rm 1}\thanks{indicates co-corresponding author.}, 
    Yi-Ping Phoebe Chen\textsuperscript{\rm 2},
    Wei Yang\textsuperscript{\rm 1}\footnotemark[1]\\
}
\affiliations{
    \textsuperscript{\rm 1}Huazhong University of Science and Technology, China\\
    \textsuperscript{\rm 2}La Trobe University, Australia\\
    \{skyesong, lr\_8823, yjqing, weiyangcs\}@hust.edu.cn, phoebe.chen@latrobe.edu.au
}

\begin{document}
% \linenumbers
\maketitle

\begin{abstract}

Transformer framework has been showing superior performances in visual object tracking for its great strength in information aggregation across the template and search image with the well-known attention mechanism. Most recent advances focus on exploring attention mechanism variants for better information aggregation. We find these schemes are equivalent to or even just a subset of the basic self-attention mechanism. In this paper, we prove that the vanilla self-attention structure is sufficient for information aggregation, and structural adaption is unnecessary. The key is not the attention structure,  but how to extract the discriminative feature for tracking and enhance the communication between the target and search image.
%In this paper, we first demonstrate that self-information enhancement in attention plays a greater role than cross-information aggregation through comprehensive theoretical analysis. %, the cross-information aggregation is indispensable in tracking but not beneficial in more.
%
%the global information exchange of the attention mechanism, which makes it natural to solve the most important problem of visual tracking: the information aggregation across the template and search image.
%
%Existing transformer-based approaches have made many variants in attention mechanism, but we found that they are essentially equivalent or even just a subset to basic attention.
%
%In this work, we figure out the component that really makes a difference in the attention mechanism, and present a masked image modeling strategy for transformer tracking. 
%
%By comprehensive theoretical analysis and experimental validations, we prove that self-information enhancement in multi image attention plays a greater role than cross-information aggregation, the cross-information aggregation is indispensable in tracking but not beneficial in more.
%
Based on this finding, we adopt the basic vision transformer (ViT) architecture as our main tracker and concatenate the template and search image for feature embedding. To guide the encoder to capture the invariant feature for tracking, we attach a lightweight correlative masked decoder which reconstructs the original template and search image from the corresponding masked tokens. The correlative masked decoder serves as a plugin for the compact transform tracker and is skipped in inference. Our compact tracker uses the most simple structure which only consists of a ViT backbone and a box head, and can run at 40 \emph{fps}.
%
%we develop a scalable masked image modeling learner, which consists of a lightweight decoder attached to the general transformer tracking framework to reconstruct the original pixels of both template and search image, and we successfully train a simple transformer tracker driven by original vision transformer architecture with significant performance gain.
%To highlight self-information in transformer tracker,
%
Extensive experiments show the proposed compact transform tracker outperforms existing approaches, including advanced attention variants, and demonstrates the sufficiency of self-attention in tracking tasks. Our method achieves state-of-the-art performance on five challenging datasets, along with the VOT2020, UAV123, LaSOT, TrackingNet, and GOT-10k benchmarks. Our project is available at \url{https://github.com/HUSTDML/CTTrack}.

\end{abstract}

\section{Introduction}

\begin{figure}[t]
\centering
\includegraphics[width=0.98\columnwidth,keepaspectratio,page=1]{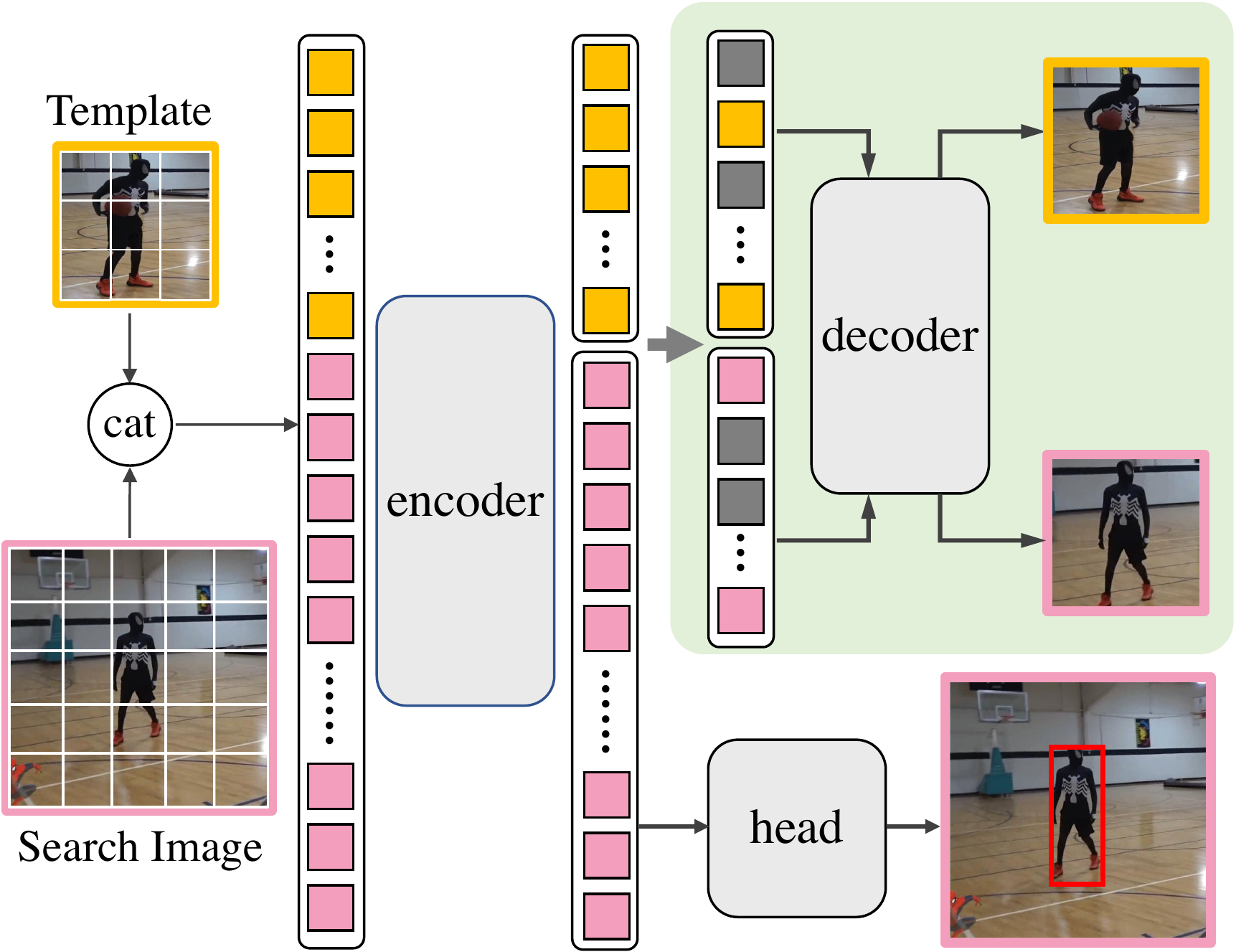}
\caption{Our compact transformer tracker adopts the simple ViT structure (encoder) with the concatenation of the template and search image as input, which essentially exploits the standard self-attention mechanism for information aggregation. The encoded tokens pass through a box head to estimate the result bounding box. And we develop a correlative masked decoder reconstructing the original template and search pixels to enhance the information aggregation, which is skipped during inference.}
\label{fig1}
\end{figure}

Visual Object Tracking is one of the fundamental tasks in computer vision with applications ranging from human-computer interaction, surveillance, traffic flow monitoring and etc. It aims to estimate the location, denoted as a bounding box, of an arbitrary target object throughout the subsequent video sequence. 
Deep Learning based trackers have achieved great success due to their strong representation ability. Trackers \cite{SiamFC, MDNet, SiamRPN, SiamRPN++} derived from Convolutional Neural Networks (CNN) \cite{AlexNet,VGG,ResNet} produce tracking accuracy that beyond the comparison of traditional approaches, especially the trackers built on Siamese network \cite{SiamFC, siamfc++, SiamRPN, SiamRPN++, SiamRCNN, SiamAttn, SiamGAT}. The key of Siamese network trackers is to produce the cross-correlation and measure the similarity between the target template and search image. Nowadays, transformer-based trackers \cite{TransT, TMT, STARK, UTT, CSWinTT, MixFormer} have shown great strength by introducing the attention mechanism~\cite{Transformer} to enhance and fuse the features of querying sample and tracked objects.
Prevalent transformer trackers~\cite{TransT, STARK, MixFormer} more or less adapt the attention for aggregating information across the template and search image.
%
% Though such a strategy is very effective and has demonstrated its outstanding performance, it xxxx (disadvantages).

%We observe that the attention mechanism, which is the most important part of the transformer, is not substantially improved, most other attention methods are equivalent variations of it or simply a subset of it. 
We find that the advanced variants of attention mechanism in recent research, including mix-attention~\cite{MixFormer} and cross-attention~\cite{SiamAttn, TransT}, are equivalent or even just a subset of the packed self-attention (i.e., standard self-attention with the concatenation of the template and search image as input). Then the question is which parts of the self-attention mechanism play an important role in visual object tracking? 
We revisited the transformer tracking framework and find that the tracking results are generated from tokens corresponding to the search image (search tokens), while the tokens corresponding to the template (template tokens) are always discarded in the last. The representational ability of search tokens comes from two parts: the cross-information enhancement from the template tokens and the self-information enhancement from the search tokens themselves. In this paper, we prove that self-information enhancement in multi-image attention plays a greater role than cross-information aggregation, though cross-information aggregation is indispensable in visual object tracking but not greatly beneficial. 
%
%Since tracking task involves two images, we define the \textbf{basic attention} in tracking field as the standard self-attention with the input of the concatenation of the template and the search image, which is widely used in previous transformer trackers\cite{STARK, CSWinTT}. 

%
%The symmetric mix attention\cite{MixFormer} is computationally equivalent to basic attention, while asymmetric mix attention and cross attention\cite{TransT, SiamAttn} are a subset of basic attention. So what exactly does basic attention consist of, and which parts really play an important role in the visual object tracking?
%

%We consider that the attention variants work well because the equivalent basic attention has a global receptive field, which grant the ability to evenly focus on information of all patches, including the template patches and search image patches, and thus implicitly strengthens the information integration.
%
Driven by this analysis, we propose a compact transformer tracker combined with correlative masked modeling for the cross-information aggregation and self-information reinforcement. As shown in Figure \ref{fig1}, our tracker adopts the basic vision transformer as the main branch and applies a lightweight masked decoder to enhance the implicit representation capability of the packed self-attention. The correlative masked decoder, which is inspired by Masked Image Modeling~\cite{MAE, SimMIM}, reconstructs the both original template and search pixels from the corresponding masked tokens, to guide the encoder to capture the invariant feature for tracking.
%reconstruct the original pixels, the encoder operates on the concatenated image patches, the decoder exploits both template tokens and search image tokens to improve the implicit representation capability in basic attention.
%Moreover, during The masked template tokens reconstruct the original template image, the masked search tokens reconstruct the template image and search image to enhance the self-information and cross-information aggregation. 
In addition, our decoder can be plugged into other transformer trackers, which can effectively improve the tracking performance without compromising speed. 
%Our masked image modeling tracking method is an extensible general model that can be easily applied to other transformer trackers, which can effectively improve the tracking performance without speed reduction. 
Applying our correlative masked modeling strategy to the compact transformer tracker can improve the AUC from 64.0\% to 65.8\% on the LaSOT~\cite{LaSOT} dataset. Extensive comparison experiments on 5 challenging datasets including VOT2020~\cite{VOT2020}, UAV123~\cite{UAV123}, LaSOT, GOT-10k~\cite{GOT-10k}, and TrackingNet~\cite{ TrackingNet} exhibits the state-of-the-art performance, which further evidence the correctness of our analysis regarding the self-attention in visual tracking. 

To summarize, our main contributions include:
\begin{enumerate}
\item We present a unified analyzing method for the attention mechanism and find that the advanced variants of the attention mechanism are equivalent or even just a subset of the self-attention. We also prove that self-information enhancement in multi-image attention plays a greater role than cross-information aggregation.

\item We develop a compact transformer tracker with a correlative masked decoder, which has a very simple structure and achieves state-of-the-art accuracy at a high Frames-Per-Seconds (\emph{fps}) tracking speed. The decoder reconstructs the original template and search image from the corresponding masked tokens and serves as a training plugin for the tracker. The experiment demonstrates that our analysis regarding self-attention is correct.

\end{enumerate}

\section{Related Work}

\noindent \textbf{Traditional trackers.}
Traditional single object tracking algorithms can be roughly summarized as Correlation Filter based trackers (CF), Deep Network based trackers (DLN). CF-based trackers\cite{MOSSE, KCF, CCOT, ECO, ATOM, DiMP} exploit the convolution theorem and learn a filter in the Fourier domain that maps known target images to the desired output. DLN-based trackers refer to algorithms employing deep neural networks for the tracking process. Earlier approaches~\cite{MDNet, DATRL} treat the tracking task as a classification problem and exploit deep features for locating the target. Shortly afterwards more trackers adopt the Siamese network~\cite{SiamFC,SiamRPN,SiamRPN++} for its effectiveness in measuring similarity. The Siamese network consists of two branches, one operates on the template and the other for the search area.

Above all, these methods mainly consist of a backbone which extracts the features of search image and template separately, a similarity measuring module, and heads to predict the location and bounding box. Compared to our framework, traditional trackers have too many modules and a very complex design, we simply adapt a ViT backbone with a box head to get better tracking results.

\noindent \textbf{Transformer trackers.}
The ViT~\cite{ViT} first introduces the transformer to image recognition tasks and presents an impressive performance. Ever since, transformer has been widely applied in image classification\cite{ViT, CVT, SwinTransformer}, object detection\cite{DERT, ViTDet}, visual object tracking\cite{STARK,TransT, TMT, CSWinTT, UTT, MixFormer} and etc.
Transformer-based tracking methods have become the mainstream tracking algorithms nowadays. TransT~\cite{TransT} proposes a feature fusion network and employs an attention mechanism to combine the features of the template and search region. STARK~\cite{STARK} develops a spatial-temporal architecture based on the encoder-decoder transformer. CSWinTT~\cite{CSWinTT} proposes a transformer architecture with multi-scale cyclic shifting window attention for visual tracking, elevating the attention from pixel level to window level. MixFormer~\cite{MixFormer} constructs a compact tracking framework and designs a mixed attention module that unifies the process of feature extraction and information matching module.

Instead of designing a complex attention mechanism as in the previous tracking approaches, we compare the essential differences of attention variants(such as mix-attention and cross-attention) and find these attention variants are equivalent or even just a subset of the packed self-attention. To verify the capability of self-attention in information aggregation, we design a compact transformer tracker using the most simple pipeline which only consists of a ViT backbone and a box head, without any extra design including separate modules of feature extraction and aggregation, and multi-layer feature aggregation.

% Information is cascaded across all patches, thus implicitly strengthens the much-needed integration of tracking between the template and the search image. We applied the above concept to compacted the transformer tracker and explored the masked image modeling methods to effectively strengthen information aggregation.

\noindent \textbf{Masked image modeling (MIM).}
MIM masks an area of the original images and predicts the missing pixels, which aims to enhance the representation of models. Recently, MIM approaches(\cite{iGPT,MAE,SimMIM,MaskFeat,BEiT}) are extended to the modern vision transformers~\cite{ViT,SwinTransformer}. iGPT~\cite{iGPT} first proposes a transformer to predict unknown pixels from a sequence of low-resolution pixels. BEiT~\cite{BEiT} tokenizes the images via an additional dVAE~\cite{dVAE} network with a block-wise masking strategy. SimMIM~\cite{SimMIM} find that a moderately large masked patch size of the input image for pixel predictions makes a strong pre-text task. MAE~\cite{MAE} develops an asymmetric encoder-decoder architecture, the encoder operates on a small proportion of the visible patches, and the decoder reconstructs the original pixels. MaskFeat~\cite{MaskFeat} reconstructs the feature descriptors such as HoG~\cite{HoG} instead of pixels.

Our approach is inspired by the previous MIM method~\cite{SimMIM, MAE}, but we have to deal with two fundamental problems in the tracking framework: (1) Visual tracking is a downstream vision task that generally does not have the pre-train process to apply the MIM strategy. We develop a masked decoder to leverage the search and the template tokens to predict the original images, which is embedded as an attachment plugin in the training phase to implement an end-to-end model. (2) MIM methods reconstructing the single image do not fit the tracking framework which involves cross-aggregation of multiple images. According to the properties of packed self-attention, we design a self-decoder and a cross-decoder to reconstruct the original template and search image from the corresponding masked tokens.
As far as we know, we are the first to artfully introduce the MIM into the visual tracking field to improve the information aggregation capabilities.

\begin{figure}[t]
\centering
\includegraphics[width=0.9\columnwidth,keepaspectratio,page=2]{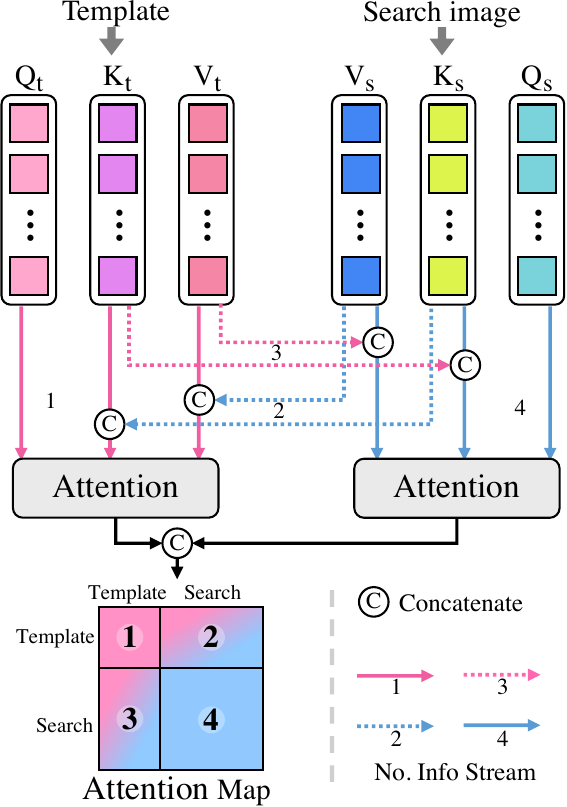}
\caption{Information streams in the attention mechanism. The four information streams of Q-K-V are corresponding to the four parts in the attention map. Variants of attention can be uniformly explained under this analytical approach.}
\label{fig2}
\end{figure}

\section{Approach}
%The main contributions of this work is designing a scalable masked image modeling (MIM) learner for cross image reconstruction which significantly improve the performance in visual object tracking. However, simply training a Siamese tracker by directly using deeper networks
%
%However, the original MIM strategy which reconstructs the single image does not fit well in a tracking framework which involves cross-aggregation of multiple images.
%
In this section, we introduce our compact transformer tracker with correlative masked modeling in detail. Before proceed, we first present a analysis on the key component of transformer tracker, and demonstrate that existing attention variants are equivalent to the packed self-attention.
%Therefore, before the introduction of the proposed MIM tracking approach, we first give a deeper analysis on the transformer tracking framework.

\subsection{Revisiting Transformer Tracker}

\noindent \textbf{Transformer tracking framework.}
As described in ViT\cite{Transformer}, the query-key-value attention mechanism is applied with query $\mathbf{Q}$, key $\mathbf{K}$, and value $\mathbf{V}$. The linear weights of $\mathbf{Q}$, $\mathbf{K}$, $\mathbf{V}$ are $\mathbf{W}_{Q}$, $\mathbf{W}_{K}$, $\mathbf{W}_{V}$ respectively. The attention (Attn) is computed as:

\begin{equation}
    \text{Attn}(\text{X}) = \text{softmax}(\dfrac{\mathbf{X}\mathbf{W}_{Q}\cdot\mathbf{W}_{K}^T\mathbf{X}^T}{\sqrt{d_k}})\cdot\mathbf{X}\mathbf{W}_{V}
\label{eq:0}
\end{equation}

\noindent where the $\text{X}$ is the input token and the $d_k$ is the dimension of the key. For a clearer description of the post-order steps, we apply an attention calculation with the inputs of two different tokens, the token $\mathbf{X}_Q$ computed with query and the token $\mathbf{X}_KV$ computed with key and value. We modify the attention formula and define the attention map (AMap) as:

\begin{equation}
\begin{split}
  \text{Attn}(\mathbf{X}_Q, \mathbf{X}_{KV}) &= \text{AMap}(\mathbf{X}_Q, \mathbf{X}_{KV})\cdot\mathbf{X}_{KV}\mathbf{W}_{V} \\
  \text{AMap}(\mathbf{X}_Q, \mathbf{X}_{KV}) &= \text{softmax}(\frac{\mathbf{X}_{Q}\mathbf{W}_{Q}\cdot\mathbf{W}_{K}^T\mathbf{X}_{KV}^T}{\sqrt{d}})
\end{split}
\label{eq:1}
\end{equation}

Our compact transformer tracker consists of two parts: a transformer backbone for information aggregation and a box head for the bounding box estimation. Give the template $z$ in the initial frame and a search image $s$. We obtain the tokens $X_t\in\mathbb{R}^{L_z\times d}$ and $X_s\in\mathbb{R}^{L_s\times d}$ respectively through patch embedding, where $d$ represents the number of channels. The \textbf{packed self-attention (PSelf-Attn)} in the tracking field is defined as the self-attention with the input of the concatenation ($\text{Cat}$) of the template and the search image:

\begin{equation}
   \text{PSelf-Attn}=\text{Attn}\Big(Cat(\mathbf{X}_z, \mathbf{X}_s), Cat(\mathbf{X}_z, \mathbf{X}_s)\Big)
\label{eq:2}
\end{equation}

\begin{figure}[t]
\centering
% \captionsetup[subfloat]{labelsep=none,format=plain,labelformat=empty}
\begin{minipage}{0.98\columnwidth}
    \centering
    \subfloat[PSelf-Attn]
    {
        \includegraphics[width=0.3\columnwidth,keepaspectratio,page=3]{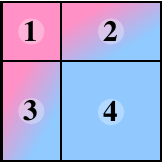}
        \label{fig:3a}
    }
    \subfloat[AMix-Attn]
    {
        \includegraphics[width=0.3\columnwidth,keepaspectratio,page=4]{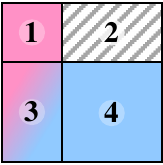}
        \label{fig:3b}
    }
    \subfloat[Cross-Attn]
    {
        \includegraphics[width=0.3\columnwidth,keepaspectratio,page=5]{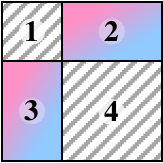}
        \label{fig:3c}
    }
    \caption{Configurations of information stream in attention map of packed self-attention (PSelf-Attn), asymmetric mix-attention(AMix-Attn) and cross-attention (Cross-Attn).}
    \label{fig:3}
\end{minipage}
\end{figure}

\noindent \textbf{Analysis on Attention.}
As shown in Figure \ref{fig2}, we divide the computation of attention mechanism, which involves both template and search image, into four information streams:
\begin{enumerate}
\setlength{\itemindent}{1.0em}
\renewcommand{\labelenumi}{(\theenumi)}
\item self-information enhancement on template;
\item cross-information aggregation on template;
\item cross-information aggregation on search image;
\item self-information enhancement on search image.
\end{enumerate}

These four information streams are also reflected in the four parts of the attention map (In Figure \ref{fig2}, the index of each part in the attention map corresponds to the information stream).
Based on this dissection, we can conveniently compare the differences between existing attention, including packed self-attention, mix-attention, and cross-attention.

The \textbf{PSelf-Attn} and the \textbf{mix-attention}\cite{MixFormer} are essentially equivalent, the mix-attention is calculated as:
\begin{equation}
\begin{split}
&\text{PSelf-Attn} == \text{Mix-Attn} = \\
\text{Cat}\Big(\text{AMap}\big(\mathbf{X}_z&, Cat(\mathbf{X}_z, \mathbf{X}_s)\big),\text{AMap}\big(\mathbf{X}_s, Cat(\mathbf{X}_z, \mathbf{X}_s)\big)\Big)\\
\end{split}
\label{eq:3}
\end{equation}
\noindent which is the same as Eqn.~\ref{eq:2}, and they include all four information streams (the attention map is shown as Figure \ref{fig:3a}). 

By the same analysis, the \textbf{asymmetric mix-attention (AMix-Attn) } contains three information streams (\#1, \#3, \#4 info stream), which is shown in the Figure \ref{fig:3b} and is calculated as follows:
\begin{equation}
\begin{split}
  &\text{AMix-Attn} = \\
  \text{Cat}\Big(\text{AMap}\big(\mathbf{X}_z, \mathbf{X}_z&\big), \text{AMap}\big(\mathbf{X}_s, Cat(\mathbf{X}_z, \mathbf{X}_s)\big)\Big)
 \end{split}
\label{eq:4}
\end{equation}

The \textbf{cross-attention} contains two information streams (\#2,\#3 info stream) for cross information aggregation, which is shown in the Figure \ref{fig:3c} and is calculated as follows:
\begin{equation}
    \text{Cross-Attn} =
    \text{Cat}\Big(\text{AMap}\big(\mathbf{X}_z, \mathbf{X}_s\big), \text{AMap}\big(\mathbf{X}_s, \mathbf{X}_z\big)\Big)
\label{eq:5}
\end{equation}

\begin{table}[htbp]\normalsize
  \centering
  \caption{The effectiveness of information streams in the attention mechanism on the LaSOT dataset. The visualized four parts in the attention map (AMap) correspond to the four information streams at the matched location.}
  \label{tab:1}
  \setlength{\tabcolsep}{2mm}{
    \begin{tabular}{c@{ }c|cccc|cc}
    \toprule
    \multirow{2}*{\#}&\multirow{2}*{AMap}&\multicolumn{4}{c|}{No. Info Stream}&\multirow{2}*{AUC}&\multirow{2}*{Prec}\\
    &&\circled{1}&\circled{2}&\circled{3}&\circled{4}&&\\
    
    \midrule
    1& \begin{minipage}[b]{0.06\columnwidth}
		\centering
		\raisebox{-.2\height}{\includegraphics[width=\linewidth,keepaspectratio,page=11]{fig.pdf}}
        \end{minipage}
    &\checkmark&\checkmark&\checkmark&\checkmark&61.7&64.2\\
    \specialrule{0em}{1pt}{1pt}
    2&\begin{minipage}[b]{0.06\columnwidth}
		\centering
		\raisebox{-.2\height}{\includegraphics[width=\linewidth,keepaspectratio,page=12]{fig.pdf}}
        \end{minipage}
    &\checkmark&&\checkmark&\checkmark&\textbf{64.0}&\textbf{67.7} \\
    \specialrule{0em}{1pt}{1pt}
    3&\begin{minipage}[b]{0.06\columnwidth}
		\centering
		\raisebox{-.2\height}{\includegraphics[width=\linewidth,keepaspectratio,page=13]{fig.pdf}}
        \end{minipage}
    &&\checkmark&\checkmark&\checkmark&60.6&63.7\\
    \specialrule{0em}{1pt}{1pt}
    4&\begin{minipage}[b]{0.06\columnwidth}
		\centering
		\raisebox{-.2\height}{\includegraphics[width=\linewidth,keepaspectratio,page=14]{fig.pdf}}
        \end{minipage}
    &\checkmark&\checkmark&\checkmark&&58.8&60.1\\
    \specialrule{0em}{1pt}{1pt}
    5&\begin{minipage}[b]{0.06\columnwidth}
		\centering
		\raisebox{-.2\height}{\includegraphics[width=\linewidth,keepaspectratio,page=15]{fig.pdf}}
        \end{minipage}
    &&\checkmark&\checkmark&& 57.9&58.5\\
    \bottomrule
    \end{tabular}
  }
\end{table}

In order to fully verify the importance of each part of packed attention, it is necessary to evaluate the impact of each information stream individually. The key of visual object tracking is to find the target in the search image, there must be a cross-information aggregation of the search image (\#3 info stream). The other information streams can be blocked out to verify their performance.

Based on the above idea, we conduct detailed experiments and the result is shown in Table~\ref{tab:1}. Removing cross-information aggregation of the template (\#2 info stream)  of self-attention can greatly improve tracking performance (the AUC and Prec of Table~\ref{tab:1} \#2 are better than that of Table~\ref{tab:1} \#1), and the cross-information aggregation of the template will introduce a lot of noise in template features, which is not recommended in visual tracking. However, removing self-information enhancement (\#3 and \#4 info stream) of self-attention severely degrades the tracking performance (the AUC and Prec of Table~\ref{tab:1} \#3 and \#4 are worse than that of Table~\ref{tab:1} \#1). From the results we can conclude that self-information enhancement in multi-image attention plays a greater role than cross-information aggregation, the cross-information aggregation is indispensable in tracking but not greatly beneficial.

% \subsection{Transformer Tracker with Correlative Masked Modeling}
\subsection{Correlative Masked Modeling}
According to the above analysis, the best tracking performance can be achieved by adopting three information streams: self-information on the template(\#1 info stream), cross-information on the search image (\#3 info stream), and self-information on the search image (\#4 info stream). These three information streams can be grouped into two categories: two self-information enhancements and one cross-information aggregation. We designed a correlative masked modeling method to enhance the information aggregation of our tracking framework, as shown in Figure \ref{fig1}. The ViT backbone is an encoder, and the correlative masked decoder reconstructs the original image (the template and search image respectively) from randomly masked tokens to enhance the self-information and reconstructs the template image from search tokens to improve cross-information aggregation. In parallel with the masked decoder, the search image tokens go through a box estimation head as in~\cite{STARK} to generate the result bounding box. 
%The decoder is embedded in the general transformer tracking framework to form an end-to-end model.

\begin{figure}[t]
\centering
\includegraphics[width=0.98\columnwidth,keepaspectratio,page=16]{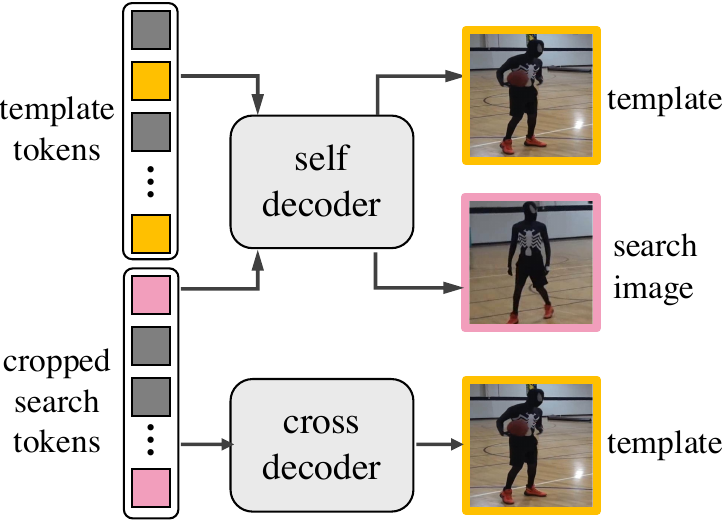}
\caption{The correlative masked decoders consists of a self-decoder and a cross-decoder. The self-decoder reconstructs the two original images, template and search image, from its corresponding masked tokens. The cross-decoder reconstructs the template image from search tokens.}
\label{fig4}
\end{figure}

\noindent \textbf{Decoder.}
The decoders in our framework consist of a self-decoder and a cross-decoder, these two decoders have the same structure but do not share weights, each one is composed of a series of transformer blocks similar to the MAE,
and the last layer of the decoder is a linear projection with output channels equal to the number of pixels in a patch.
As shown in Figure~\ref{fig4}, the decoder takes masked tokens as input and predicts the original image pixels corresponding to the template token and the search image token, where the template tokens are only self-reconstructed to the template image for enhancing the \#1 information stream, search tokens are used to crossly reconstruct the template image (for \#3 info stream) and self-reconstruct the search image (for \#4 info stream). 

\noindent \textbf{Masking and Reconstruction.}
The encoder embeds the concatenation set of template tokens and search tokens. Then we split the encoded tokens into template tokens and search tokens, crop the search tokens using Precise RoI Pooling\cite{prroipooling} to the same size as the template tokens, and sample a subset of them. We randomly sample tokens at a high masking ratio (75\%). Our decoder predicts the pixel values for each masked token, and the output of the decoder is reshaped to form a reconstructed image. We use the mean squared error (MSE) between the reconstructed and original images on masked tokens as our loss function.

\subsection{Training and Inference}
Our decoder is only used in the training phase, while does not participate in the inference phase, hence it doesn't affect the tracking speed.
During the training phase, our tracker takes a triplet input consisting of one search region and two templates similar to STARK\cite{STARK}. We randomly sample multiple frames from sequences in the training set, select the first frame and the second frame as templates, and the last frame as the search region. In the target localization training, we train the whole network except the scoring head in an end-to-end manner with the combination of $L1$ Loss, generalized IoU loss~\cite{giou}, and decoder loss $L_{dec}$. The full loss function is defined as follows:
\begin{equation}
Loss = \lambda_{L1}L_1(B_i,\hat{B}_i)+\lambda_{g}L_{g}(B_i,\hat{B}_i) + \lambda_{dec}L_{dec}
\label{eq:loss1}
\end{equation}
\noindent where $\lambda_{L1} = 5.0$, $\lambda_{g} = 2.0$ and $\lambda_{dec} = 0.3$ are the weighting factors of three losses, $\hat{B}_i$ is the estimated box of the target and $B_i$ is the ground-truth bounding box. The decoder loss $L_{dec}$ is defined as:
\begin{equation}
L_{dec} = L_2(z,z_p)+ L_2(s,s_p)+ L_2(z,s_p)
\label{eq:loss1}
\end{equation}
\noindent where the $L_2$ is the MSE loss, $z$ and $s$ represent the original template image and search image, $z_p$ and $s_p$ represent the predicting template image and search image respectively.

In the inference phase, we use two templates of the same size as the input. One of which is the initial template and fixed, the other is online updated and always set to the latest tracking result with high confidence. We use a score head to control the updating of the online template. Our score head consists of the multilayer perceptron (MLP) that receives a class-token\cite{ViT} as input and evaluates the accuracy of current tracking results. 

\section{Experiments}

\subsection{Implementation Details}
In order to effectively verify the correctness of our analysis, we design the compact transformer tracker without any other extra attention mechanisms. The only structures remaining are feature extraction and aggregation, and multi-layer feature aggregation. The main tracker only consists of a ViT backbone and a box estimation head, we test both ViT-Base and ViT-Large, and the ViT parameters are initialized with MAE \cite{MAE} pre-trained model. We refer our \textbf{C}ompact \textbf{T}ransformer tracker as CTTrack-B (the backbone of ViT-Base) and CTTrack-L (the backbone of ViT-Large) in this section.

We adopt CoCo\cite{COCO}, LaSOT\cite{LaSOT}, GOT-10k\cite{GOT-10k}, and TrackingNet\cite{TrackingNet} as our training dataset except the GOT-10k benchmark. The training samples are directly sampled from the same sequence and we apply common data augmentation operations including brightness jitter and horizontal flip. The size of the input template is 128$\times$128, the search region is $5^2$ times of the target box area and further resized to 320$\times $320. The decoder parameters are initialized with Xavier Uniform. The AdamW optimizer \cite{adamw} is employed with initial learning rate (lr) of 1e-4 with the layer-wise decay 0.75, and the lr decreases according to the cosine function with the final decrease factor of 0.1. We adopt a warm-up lr with the 0.2 warm-up factor on the first 5 epochs. We train our model on 4 Nvidia Tesla V100 GPUs for a total of 500 epochs, each epoch uses $6 \times 10^4$ images. The mini-batch size is set to 128 images with each GPU hosting 32 images. Our approach is implemented in Python 3.7 with PyTorch 1.7.

\subsection{Ablation Study}

We ablate our compact transformer tracker on several intriguing properties using the challenging LaSOT dataset and report the Area Under the Curve (AUC) and Precision (Prec) as the validation accuracy.

\begin{table}[htbp]\normalsize
  \centering
  \caption{Model size and speed using different backbones.}
  \label{tab:aba1}
  \setlength{\tabcolsep}{2.5mm}{
    \begin{tabular}{c|ccc}
    \toprule
    Methods&Params(M)&FLOPs(G)&Speed(\emph{fps})\\
    \midrule
    CTTrack-B&93.8&48.1&40\\
    CTTrack-L&313.9&163.7&22\\
    \bottomrule
    \end{tabular}
  }
\end{table}

\noindent \textbf{Backbone Comparison.}
Table \ref{tab:aba1} shows the comparison of the transformer backbones between the ViT-Base and ViT-Large backbone. The CTTrack-B reaches a higher tracking speed while the CTTrack-L exhibits a better performance.

\noindent \textbf{Reconstruction Streams.}
Our decoder enforces three types of reconstruction streams as shown in Figure \ref{fig4}. Table \ref{tab:aba2} exhibits different configurations of reconstruction streams, through varied combinations of search tokens reconstruct search image (s2s), template tokens reconstruct template image (t2t) and search tokens reconstruct template image(s2t). The result is consistent with the conclusion of our previous analysis that self-information enhancement (\#5) plays the most important role in transformer tracking, compared to cross-information aggregation(\#4). Besides, search image information has more influence than the template information, the s2s (\#2) improves performance the most among all streams (\#2, \#3, \#4), from 64.0 to 64.7 in AUC score. After adopting all three reconstruction streams, tracking accuracy improved by an impressive AUC score of 1.8\%, which validates the effectiveness of our masked modeling decoders.

\begin{table}[htbp]\normalsize
  \centering
  \caption{Ablation Study for the reconstruction streams. \textbf{s2s} represents search tokens reconstruct search image, \textbf{t2t} denotes template tokens reconstruct template image and \textbf{s2t} means search tokens reconstruct template image.}
  \label{tab:aba2}
  \setlength{\tabcolsep}{3mm}{
    \begin{tabular}{c|@{}ccccc@{}|cc}
    \toprule
    \multirow{2}*{\#}&&\multicolumn{3}{c}{Recons Type}&&\multirow{2}*{AUC}&\multirow{2}*{Prec}\\
    \cline{3-5}
    &&s2s&t2t&s2t&&&\\
    \midrule
    1&&-&-&-&&64.0&67.7\\
    2&&\checkmark&-&-&&64.7&69.1\\
    3&&-&\checkmark&-&&64.4&68.4\\
    4&&-&-&\checkmark&&64.4&68.6\\
    5&&\checkmark&\checkmark&-&&65.1&69.9\\
    6&&\checkmark&\checkmark&\checkmark&&\textbf{65.8}&\textbf{70.9}\\
    \bottomrule
    \end{tabular}
  }
\end{table}

\noindent \textbf{Masking ratio.}
When we conduct reconstruction streams, we randomly mask the input tokens according to a pre-defined ratio. Table \ref{tab:aba3} shows the influence of different masking ratios. We mask the encoded template token and search tokens with a random sampling strategy at different masking rates. Similar to the conclusion obtained by the MAE\cite{MAE},  the optimal ratios are relatively high, and the accuracy increases steadily with the masking ratio growing until reaching 75\%,  which produces the best tracking results. 

\begin{table}[htbp]\normalsize
  \centering
  \caption{Comparison on masking ratio.}
  \label{tab:aba3}
  \setlength{\tabcolsep}{2mm}{
    \begin{tabular}{c|cccc}
    \toprule
    Mask Ratio&25\%&50\%&75\%&90\%\\
    \midrule
    AUC&64.6&65.7&\textbf{65.8}&64.9\\
    Prec&69.0&70.7&\textbf{70.9}&69.5\\
    \bottomrule
    \end{tabular}
  }
\end{table}

\noindent \textbf{Online Template Updating.}
We evaluate the effect of the online update strategy in our method. The ablation study result is shown in Table \ref{tab:aba4}, \#1 represents the performance without template updating. We can see that applying a fixed interval to update the online template (\#2) is ineffective as it greatly reduces the quality of template and causes tracking drift. It can be seen in \#3, there is a 0.2\% improvement in the AUC score after applying the scoring head to evaluate the accuracy of current tracking results.

\begin{table}[htbp]\normalsize
  \centering
  \caption{Ablation for the online template updating component.\textbf{Online} denotes updating the template at a fixed update interval. \textbf{Score} represents the online template is only updated with high confident samples. }
  \label{tab:aba4}
  \setlength{\tabcolsep}{2mm}{
    \begin{tabular}{c|cc|cc}
    \toprule
    &Online&Score&AUC&Prec\\
    \midrule
    \multirow{3}*{CTTrack-B}&-&-&65.8&70.9\\
    &\checkmark&-&64.9&69.9\\
    &\checkmark&\checkmark&66.0&71.1\\
    \bottomrule
    \end{tabular}
  }
\end{table}

\begin{figure}[t]
\centering
\includegraphics[width=0.98\columnwidth,keepaspectratio,page=17]{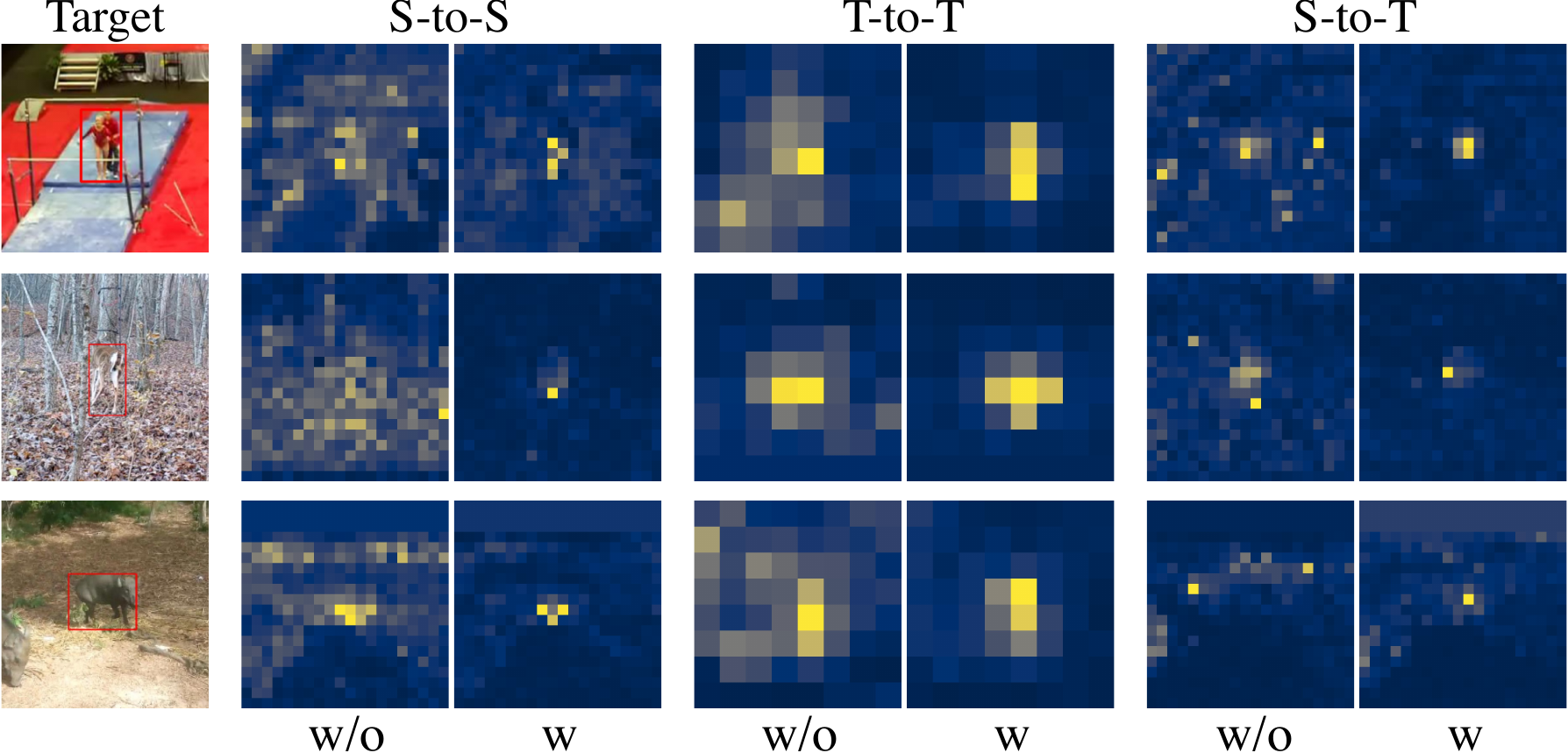}
\caption{Visualization of attention map which compares the difference between training with correlative decoder (w) and training without correlative decoder(w/o). \textbf{S-to-S} is self-information enhancement on search image, \textbf{T-to-T} is self-information enhancement on template, \textbf{S-to-T} is cross-information aggregation on search image.}
\label{fig:visual}
\end{figure}

\noindent \textbf{Visualization of attention maps.}
We visualize attention maps in Figure\ref{fig:visual}, our tracker adopting the correlative decoder has a stronger discriminative ability. The baseline transformer without a reconstruction decoder tends to lose the target position, and the distractors in the background get suppressed with the training by the correlative decoder.

\subsection{Comparison with the SOTA}

\begin{table*}[htbp]\small
  \centering
  \caption{Comparisons with previous state-of-the-art trackers on four challenge benchmarks. The \textcolor{red}{\textbf{red}}, \textcolor{cGreen}{green} and \textcolor{blue}{blue} indicate performances ranked at first, second, and third places. The \textbf{tracker -GOT} denotes only trained on the GOT-10k train split.}
  \label{tab:main1}
  \setlength{\tabcolsep}{2mm}{  
  \begin{tabular}{l cc c ccc c ccc c ccc}
    \toprule
    \multirow{2}*{Methods} & \multicolumn{2}{c}{UAV123}&& \multicolumn{3}{c}{LaSOT} && \multicolumn{3}{c}{TrackingNet} && \multicolumn{3}{c}{GOT-10k}\\
    \cline{2-3}
    \cline{5-7}
    \cline{9-11}
    \cline{13-15}
    & AUC&P && AUC&P$_{Norm}$&P && AUC&P$_{Norm}$&P && AO&SR$_{0.5}$&SR$_{0.75}$\\
    \midrule[0.5pt]
    \textbf{CTTrack-L}& \textcolor{red}{\textbf{71.3}}&\textcolor{red}{\textbf{93.3}} && \textcolor{cGreen}{69.8}&\textcolor{cGreen}{79.7}&\textcolor{cGreen}{76.2} && \textcolor{red}{\textbf{84.9}}&\textcolor{red}{\textbf{89.1}}&\textcolor{red}{\textbf{83.5}} && 75.3&84.5&74.0 \\
    \textbf{CTTrack-B}& 68.8&89.5 && \textcolor{blue}{67.8}&\textcolor{blue}{77.8}&\textcolor{blue}{74.0} && \textcolor{blue}{82.5}&\textcolor{blue}{87.1}&\textcolor{blue}{80.3} && 73.5&83.5&70.6 \\
    \textbf{CTTrack-L -GOT}& -&- && -&-&- && -&-&- &&
    \textcolor{red}{\textbf{72.8}}&\textcolor{red}{\textbf{81.3}}&\textcolor{red}{\textbf{71.5}} \\
    \textbf{CTTrack-B -GOT}& -&- && -&-&- && -&-&- && \textcolor{cGreen}{71.3}&\textcolor{cGreen}{80.7}&\textcolor{cGreen}{70.3} \\
    \hline
    MixFormer\cite{MixFormer}& \textcolor{blue}{69.5}&\textcolor{cGreen}{91.0} && \textcolor{red}{\textbf{70.1}}&\textcolor{red}{\textbf{79.9}}&\textcolor{red}{\textbf{76.3}} && \textcolor{cGreen}{83.9}&\textcolor{cGreen}{88.9}&\textcolor{cGreen}{83.1} && \textcolor{blue}{70.7}&80.0&\textcolor{blue}{67.8} \\
    CSWinTT\cite{CSWinTT}& \textcolor{cGreen}{70.5}&\textcolor{blue}{90.3}&& 66.2&75.2&70.9 && 81.9&86.7&79.5 && 69.4&78.9&65.4  \\
    UTT\cite{UTT}& - & - &&  64.6 & - & 67.2 && 79.7 & - & 77.0 && 67.2 & 76.3 & 60.5 \\
    STARK\cite{STARK}& -&- &&67.1&77.0&- && 82.0&86.9&- && 68.8&78.1&64.1  \\
    TransT\cite{TransT}& 68.1& 87.6 && 64.9&73.8&69.0 && 81.4&86.7&\textcolor{blue}{80.3} && 67.1&76.8& 60.9 \\
    TrDiMP\cite{TMT}& 67.0&87.6 && 64.0&73.2&66.6 && 78.4&83.3&73.1 && 68.8&\textcolor{blue}{80.5}&59.7 \\
    STMTrack\cite{stmtrack}& 64.7 & - && 60.6 & 69.3 & 63.3 && 80.3 & 85.1 & 76.7 && 64.2 & 73.7 & 57.5 \\
    AutoMatch\cite{AutoMatch}& 64.4&83.8 && 58.2&67.5& 59.9 && 76.0&82.4&72.5 && 65.2&76.6&54.3 \\
    SiamGAT\cite{SiamGAT}& 64.6&84.3 && 53.9&63.3&53.0 && -&-&- && 62.7&74.3&48.8 \\
    % SiamRCNN\cite{SiamRCNN}& -&- && 64.8&72.2&- && 81.2&85.4&\textcolor{blue}{80.0} && 64.9&72.8&59.7 \\
    % Ocean\cite{OCEAN}& 62.1&82.3 && 51.6&60.7&52.6 && 69.2&79.4&68.7 && 61.1&72.1&47.3 \\
    % PrDiMP\cite{PrDiMP}& 66.6&87.2 && 59.9&68.8&60.8 && 75.8&81.6&70.4 && 63.4&73.8&54.3 \\
    KYS\cite{KYS}& -&- && 55.4&63.3&55.8 && 74.0&80.0&68.8 && 63.6&75.1&51.5 \\
    % D3S\cite{D3S}& -&- && -&-&- && 72.8&76.8&66.4 && 59.7&67.6&46.2 \\
    MAML\cite{MAML}& -&- && 52.3&-&53.1 && 75.7&82.2&72.5 && -&-&- \\
    SiamAttn\cite{SiamAttn}& 65.0&84.5 && 56.0&64.8&- && 75.2&81.7&- && -&-&- \\
    SiamFC++\cite{siamfc++}& 61.8&80.4 && 54.4&62.3&54.7 && 75.4&80.0&70.5 && 59.5&69.5&47.9 \\
    SiamRPN++\cite{SiamRPN++}& 64.2&84.0 && 49.6&56.9&49.1 && 73.3&80.0&69.4 && 51.7&61.6&32.5 \\
    DiMP\cite{DiMP}& 64.2&84.9 && 57.7&66.4&57.9 && 74.0&80.1&68.7 && 61.1&71.7&49.2 \\
    ATOM\cite{ATOM}& 61.7&82.7 && 51.5&57.6&50.5 && 70.3&77.1&64.8 && 55.6&63.4&40.2 \\
    % ECO\cite{ECO}& 52.5&74.1 && 32.4&33.8&30.1 && 55.4&61.8&49.2 && 31.6&30.9&11.1 \\
    % SiamFC\cite{SiamFC}& 49.2&72.7 && 33.6&42.0&33.9 && 57.1&66.3&53.3 && 34.8&35.3&9.8 \\
    % TrTr\cite{TrTr}& 1&2 && 1&2&3 && 1&2&3 && 1&2&3  \\
    % STARK100\cite{STARK}& \textcolor{cGreen}{69.2}&\textcolor{cGreen}{88.2} && \textcolor{cGreen}{67.1}&\textcolor{cGreen}{77.0}&\textcolor{cGreen}{-} && \textcolor{blue}{82.0}&\textcolor{blue}{86.9}&- && \textcolor{blue}{68.8}&78.1&\textcolor{blue}{64.1} \\
  \bottomrule
\end{tabular}
}
\end{table*}

\begin{table}[htbp]\normalsize
  \centering
  \caption{Comparisons on VOT2020, where trackers only predict bounding boxes rather than masks.}
  \label{tab:main2}
  \setlength{\tabcolsep}{2mm}{
    \begin{tabular}{l|ccc}
    \toprule
    % &\multicolumn{3}{c}{VOT2020\cite{VOT2020}}\\
    % \cline{2-4}
    Methods&EAO$\uparrow$&Accuracy$\uparrow$&Robustness$\uparrow$\\
    \midrule[0.5pt]
    SiamFC&0.179&0.418&0.502 \\
    ATOM&0.271&0.462&0.734 \\
    DiMP&0.274&0.457&0.740 \\
    UPDT&0.278&\textcolor{blue}{0.465}&\textcolor{blue}{0.755} \\
    TransT& \textcolor{cGreen}{0.293}&\textcolor{cGreen}{0.477}&0.754 \\
    % TrTr\cite{TrTr}& 1&2&3 \\
    % STARKST50& \textcolor{cGreen}{0.303}&\textcolor{red}{\textbf{0.481}}&\textcolor{blue}{0.775} \\
    CSWinTT& \textcolor{red}{\textbf{0.304}}&\textcolor{red}{\textbf{0.480}}&\textcolor{red}{\textbf{0.787}} \\
    \midrule[0.1pt]
    CTTrack-L& \textcolor{blue}{0.287}&0.453&\textcolor{red}{\textbf{0.787}} \\
    \bottomrule
    \end{tabular}
  }
\end{table}

We compare our compact tracker with the state-of-the-art trackers on UAV123\cite{UAV123}, LaSOT\cite{LaSOT}, TrackingNet\cite{TrackingNet}, GOT-10k\cite{GOT-10k}, and VOT2020\cite{VOT2020}. For a fairer comparison, here we adopt relative position biases in our ViT backbones, this addition improves AUC by around 1 point.

\noindent \textbf{UAV123} gathers an application-specific collection of 123 sequences. It adopts the AUC and Precision (P) as the evaluation metrics. As shown in Table \ref{tab:1}, Our CTTrack-L outperforms previous trackers and exhibits very competitive performance (71.3\% AUC) when compared to the previous best-performing tracker CSWinTT (70.5\% AUC).

\noindent \textbf{LaSOT} is a long-term dataset including 1400 sequences and distributed over 14 attributes, the testing subset of LaSOT contains 280 sequences. Methods are ranked by the AUC, P, and Normalized Precision (P$_{Norm}$). Our CTTrack-L achieves the AUC (69.8\%) and Prec (76.2\%), which is an excellent result that outperforms other methods only except the MixFormer. Our tracker has lower performance than MixFormer on LaSOT because it contains long-term sequences and large variations in content. ViT backbone is a plain and non-hierarchical architecture that maintains feature maps at a certain scale, which may not be able to well handle long-term tracking sequences with scale variations.

\noindent \textbf{TrackingNet} is a large-scale tracking dataset consisting of 511 sequences for testing. 
The evaluation is performed on the online server. 
Table \ref{tab:1} shows that CTTrack-L performs better quality and ranks first in AUC score at 84.9\%. The gain is 1.0\% improvement when compared with the previous best results.

\noindent \textbf{GOT-10k} contains over 10k videos for training and 180 for testing. It forbids the trackers to use external datasets for training. We follow this protocol by retraining our trackers to only use the GOT10k train split. As in Table \ref{tab:1}, MixFormer and CSWinTT provide the best performance, with an AO score of 70.7\% and 69.4\%. Our CTTrack-L has obtained an AO score of 72.8\%, significantly outperforming the best existing tracker by 2.1\%.

\noindent \textbf{VOT2020} benchmark contains 60 challenging videos. The performance is evaluated using the expected average overlap (EAO), which takes both accuracy (A) and robustness (R). 
% In addition, a new anchor-based evaluation protocol is proposed in VOT2020, the segmentation mask is adopted as the ground truth. However, 
Since our algorithm does not output a segmentation mask, trackers that only predict bounding boxes are selected for comparisons to ensure fairness. It can be seen from Table \ref{tab:main2} that our CTTrack-L obtains an EAO of 0.287.

\section{Conclusion}

In this work, we analyze the information stream in the attention mechanism in depth. We prove that the vanilla self-attention structure is sufficient for information aggregation, and employ the three information streams of the packed self-attention in the transformer tracking framework.
%and present a correlative masked modeling strategy for a compact transformer tracker. 
%
To enhance the information representation, we design the correlative masked decoder consisting of a self-decoder and a cross-decoder to reconstruct the original pixels of both template and search image.
Extensive experiments demonstrate the effectiveness of our correlative masked modeling strategy and our compact transformer tracker exhibits impressive performance over previous trackers. 
In addition, our correlative masked decoder can be plugged into other transformer trackers, which can effectively improve the tracking performance without compromising speed.
In the future, we plan to combine the feature pyramid or convolution module for better performance on long-term tracking sequences. 

% we consider extending our correlative masked modeling to multiple object tracking.

\section*{Acknowledgments}
This work is supported by the national key research and development program of China under Grant No.2020YFB1805601, National Natural Science Foundation of China (NSFC No. 62272184), and CCF-Tencent Open Research Fund (CCF-Tencent RAGR20220120). The computation is completed in the HPC Platform of Huazhong University of Science and Technology.

\bibliography{aaai23}
% \nobibliography{aaai23}
\bigskip

\end{document}